\pdfoutput=1

\documentclass[11pt]{article}

\usepackage[final]{acl}

\usepackage{times}
\usepackage{latexsym}

\usepackage[T1]{fontenc}

\usepackage[utf8]{inputenc}

\usepackage{microtype}

\usepackage{inconsolata}

\usepackage{graphicx}
\usepackage{amsmath}  
\usepackage{bm}
\usepackage{graphicx}
\usepackage{xcolor}
\usepackage{multirow}
\usepackage{tabularx}
\usepackage{booktabs}
\usepackage{pgfplots}
\usepackage{tcolorbox}
\pgfplotsset{compat=1.18}

\title{DuplexMamba: Enhancing Real-time Speech Conversations with Duplex and Streaming Capabilities}

\author{Xiangyu Lu$^{1}$, Wang Xu$^{2*}$, Haoyu Wang$^{2}$, Hongyun Zhou$^{1}$, Haiyan Zhao$^{2}$, \\
\textbf{Conghui Zhu$^{1}$\thanks{Corresponding authors: Wang Xu and Conghui Zhu}, Tiejun Zhao$^{1}$, Muyun Yang$^{1}$}\\
{$^{1}$Faculty of Computing, Harbin Institute of Technology, Harbin, China}\\
{$^{2}$Tsinghua University, Beijing, China} \\
\texttt{\{lu9995801,xwjim812\}@gmail.com}, \texttt{conghui@hit.edu.cn}
}

\begin{document}
\maketitle
\begin{abstract}
Real-time speech conversation is essential for natural and efficient human-machine interactions, requiring duplex and streaming capabilities.
Traditional Transformer-based conversational chatbots operate in a turn-based manner and exhibit quadratic computational complexity that grows as the input size increases.
In this paper, we propose DuplexMamba, a Mamba-based end-to-end multimodal duplex model for speech-to-text conversation.
DuplexMamba enables simultaneous input processing and output generation, dynamically adjusting to support real-time streaming.
Specifically, we develop a Mamba-based speech encoder and adapt it with a Mamba-based language model.
Furthermore, we introduce a novel duplex decoding strategy that enables DuplexMamba to process input and generate output simultaneously.
Experimental results demonstrate that DuplexMamba successfully implements duplex and streaming capabilities while achieving performance comparable to several recently developed Transformer-based models in automatic speech recognition (ASR) tasks and voice assistant benchmark evaluations.
Our code and model are released~\footnote{Code and model: \url{https://github.com/khfs/DuplexMamba.git}}.
\end{abstract}

\section{Introduction}
\label{sec1}
Large language models (LLMs) have transformed human-machine interactions, showcasing exceptional capabilities in diverse applications such as daily assistance~\cite{2022chatgpt,achiam2023gpt} and task automation~\cite{wangvoyager,qian2024chatdev,2024gpt4o}. 
As artificial intelligence systems become increasingly integrated into daily life, the ability to conduct streaming real-time conversations has emerged as a critical challenge in human-machine interaction. 
Recent efforts have focused on improving the interactive capabilities of LLMs, including duplex~\cite{zhang2024beyond,fu2024vita} and streaming capabilities~\cite{defossez2024moshi,yao2024minicpm}. 

Traditional audio-language models depend on a cascaded paradigm~\cite{huang2024audiogpt,shen2024hugginggpt}, where ASR models and LLMs operate in sequential connection. 
These cascaded systems, constructed from discrete modules, suffer from error propagation during execution and present significant challenges for unified system optimization.
To address the limitations of the cascaded paradigm, researchers have developed various end-to-end audio LLMs~\cite{chu2024qwen2,tangsalmonn} that integrate speech encoders with LLMs through speech adapters~\cite{fang2024llama,xie2024mini} or multilayer perceptron (MLP) layers~\cite{fu2024vita}, enabling comprehensive end-to-end optimization of speech processing. 
However, these models are primarily based on Transformer~\cite{vaswani2017attention} architectures, which utilize attention mechanisms that scale quadratically with sequence length, resulting in prohibitive computational costs for long conversations.

Duplex capability, which denotes simultaneous input processing and output generation, is essential for real-time interaction.
The capability remains notably absent in current turn-based language models, and Various attempts have been made to develop duplex models: 
MiniCPM-duplex~\cite{zhang2024beyond} implements time-division multiplexing by dividing queries and responses into time slices for pseudo-simultaneous processing. 
LSLM~\cite{ma2024language} enables real-time turn-taking detection by combining input and output tokens for autoregressive generation. 
Moshi~\cite{defossez2024moshi} achieves parallelism by simultaneously modeling both input and output speech streams.

This paper proposes DuplexMamba, a novel end-to-end multimodal duplex model for speech-to-text conversations built on the Mamba~\cite{gu2023mamba} architecture. 
Specifically, we develop a Mamba-based speech encoder and adapt it with a Mamba-based language model.
The development process involves two training stages: multimodal alignment and multimodal instruction tuning. 
To achieve duplex and streaming capabilities, we introduce an innovative duplex strategy that incorporates state tokens to indicate input states. 
Through two additional training stages: input state discrimination and streaming alignment, our model can effectively predict state tokens and process input streamingly.
A distinguishing feature of the Mamba architecture is its fixed-size contextual memory, which results in linear computational and memory complexity per token relative to sequence length during inference. 
This characteristic is fundamental to our model's streaming capability. 

Experimental results demonstrate that DuplexMamba successfully implements duplex and streaming capabilities while achieving performance comparable to several recently developed Transformer-based models across ASR tasks and voice assistant benchmark evaluations.

\begin{figure*}[t!]
    \centering
    \includegraphics[width=\linewidth]{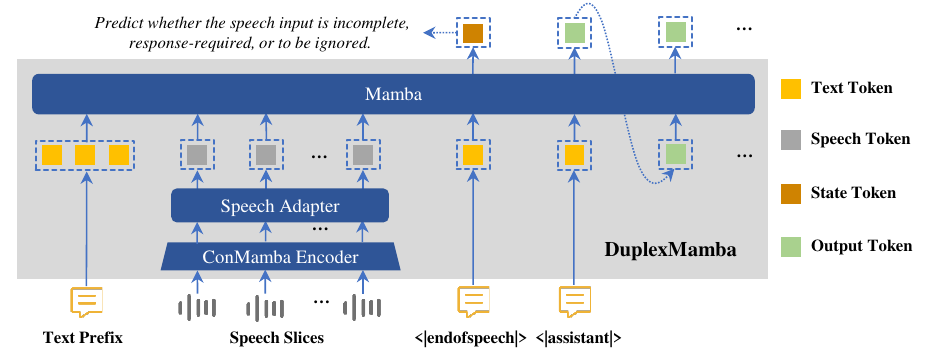}
    \caption{The model architecture of DuplexMamba.}
    \label{fig:architecture}
\end{figure*}

\section{Preliminary}
\label{sec2}
\subsection{Mamba}
Mamba is a novel neural network architecture introduced in ~\citet{gu2023mamba}, marking an advancement over the traditional Transformer architecture.
Mamba employs the selective state space model (SSM) to replace the self-attention mechanism used in Transformers.
SSM dynamically adjusts state transitions and observation processes, enabling the model to better capture key information in sequences.
The SSM is shown in Equation~\ref{eq:ssm}:
\begin{equation}
\bm{h}_t = \bm{\overline{A}} \bm{h}_{t-1} + \bm{\overline{B}} \bm{x}_t , \quad \bm{y}_t = \bm{C} \bm{h}_t
\label{eq:ssm}
\end{equation}
where $t$ represents the current time step, and $\bm{h}_t$ is the state or recurrent state, $\bm{x}_t$ and $\bm{y}_t$ represent the input and output at time $t$, respectively.
In Mamba's selective SSM implementation, the matrices $\bm{\overline{A}}$, $\bm{\overline{B}}$, and $\bm{C}$ are not static but dynamically computed based on the input $\bm{x}_t$.
Unlike the KV cache in Transformer models, $\bm{h}_t$ has a fixed size and does not grow with the context length. 

\subsection{ConMamba}
\label{sec2.1}
ConMamba is first introduced in ~\citet{jiang2024speech} to enhance Mamba's performance on ASR tasks. 
While Mamba is inherently unidirectional and causal, speech processing tasks typically benefit from bidirectional modeling that integrates both past and future contextual information. 
To address non-causal tasks, Bidirectional Mamba is proposed in \citet{zhuvision}. 
This architecture runs two-directional SSMs and causal convolutions. %
The outputs from both directions are averaged to incorporate information from both temporal perspectives. 
The ConMamba block comprises three main components: bidirectional Mamba, feedforward layers, and convolutional modules. 

\section{DuplexMamba}
\subsection{Model Architecture}
\label{sec3}
In this section, we introduce the model architecture of DuplexMamba. 
As shown in Figure~\ref{fig:architecture}, it consists of a ConMamba speech encoder, a speech adapter, and a Mamba language model.

\paragraph{Speech Encoder}
First, we train a Mamba-based ASR model following~\citet{jiang2024speech}, whose architecture integrates a ConMamba encoder with a Mamba decoder. 
The ConMamba encoder comprises multiple stacked ConMamba blocks positioned downstream of a CNN~\cite{lecun1998gradient} frontend, which compresses the input spectrogram into tokens. 
As demonstrated in~\citet{jiang2024speech}, this model achieves performance comparable to similarly-sized Transformer-based models.

Then we use the ConMamba encoder of the trained ASR model as our speech encoder, denoted as $\mathrm{E}$.
Specifically, for a user’s speech input $\bm{X}^S$, the encoded speech representation is given as $\bm{H}=\mathrm{E}(\bm{X}^S)$, where $\bm{H}$ is the sequence of speech representation.

\paragraph{Speech Adapter}
We introduce a speech adapter, between the speech encoder and the language model. 
This adapter bridges the audio-text modality gap by mapping speech representations into the embedding space of the language model.
Following~\citet{ma2024embarrassingly} and \citet{fang2024llama}, our adapter first downsamples the speech representations $\bm{H}$ to reduce the sequence length, concatenating every $k$ consecutive frames along the feature dimension to obtain $\bm{H}'$.
Then $\bm{H}'$ is passed through a two-layer perceptron with a ReLU activation between the linear layers to produce the final speech representation $\bm{S}$. 

\paragraph{Language Model}
As shown in Figure~\ref{fig.template}, the resulting speech representation sequence is concatenated with the representation of text tokens based on a prompt template $\mathcal{T}$, which is then fed into the Mamba-based language model.
$\bm{S}$ denotes the speech representation sequence. 
The complete sequence, denoted as $\mathcal{T}(\bm{S})$, is fed into the language model, which autoregressively generates the text response $\bm{Y} = [\bm{y}_1, ..., \bm{y}_M]$, where $M$ represents the length of the generated sequence. 

\begin{figure*}[t!]
    \centering
    \includegraphics[width=\linewidth]{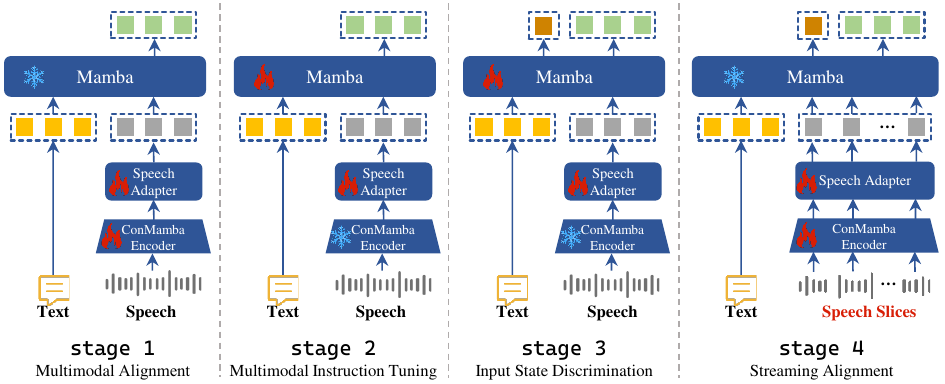}
    \caption{The four-stage training of DuplexMamba.}
    \label{fig:training}
\end{figure*}

\subsection{Training Procedure}
\label{sec4}
As illustrated in Figure~\ref{fig:training}, our training procedure consists of four stages: multimodal alignment, multimodal instruction tuning, input state discrimination, and streaming alignment.

\paragraph{Multimodal Alignment}
In this stage, we leverage the ASR task to align the representation spaces between the speech encoder and the language model.
Specifically, we construct training data from ASR datasets.
The model is designed to process speech input and generate the corresponding transcribed text.
The prompts are listed in Appendix~\ref{sec:appendix_1}.
To increase diversity, multiple prompts are generated using Chat-GPT.
 
We employ a cross-entropy loss function to guide the optimization process, formally represented by the following equation:
\begin{equation}
\mathcal{L} = - \sum_{i=1}^{M} \log P(\bm{y}_i^T \mid \mathcal{T}(\bm{S}), \bm{Y}_{<i}^T)
\label{eq:LLM_loss}
\end{equation}

During training, only the ConMamba encoder and speech adapter parameters are optimized, while the Mamba language model remains frozen.

\paragraph{Multimodal Instruction Tuning}
In this stage, we conduct instruction tuning to enhance the model's instruction-following capability across multimodal contexts. 
During training, the model receives speech input and generates a textual response by integrating both the textual instruction and speech content. 
The training tasks encompass ASR and speech-to-text QA tasks. 
We use the ASR text prompts previously employed in the multimodal alignment stage and use ChatGPT to generate seven QA text prompts as described in Appendix~\ref{sec:appendix_2}. 

The optimization process is guided by the loss function defined in Equation~\ref{eq:LLM_loss}. 
We freeze the ConMamba encoder and focus the training exclusively on the Mamba language model and speech adapter.

\paragraph{Input State Discrimination}
\label{state_discrimination}
To facilitate duplex decoding, we propose state tokens that explicitly denote the status of the input.
Specifically, we introduce three distinct state tokens:
1) \textbf{<response>}: Indicates that the input is complete and necessitates a response. 
2) \textbf{<incomplete>}: Signifies that the input is incomplete. 
3) \textbf{<ignore>}: Denotes that the input is complete but should be ignored.
The implementation and detailed mechanism of these state tokens will be elaborated in Section~\ref{branch_switching}.

We construct a state discrimination dataset and add state tokens after the <$|$endofspeech$|$> marker in prompts as shown in Figure~\ref{fig.template}. 
Specifically, the response-required data is directly extracted from QA datasets.
The incomplete data comprises speech inputs from the same dataset, wherein audio segments are randomly truncated, and the corresponding answers are replaced with one of six predefined textual labels detailed in Appendix~\ref{sec:appendix_3} to indicate an incomplete input. 
For the ignored data, we leverage the ASR dataset, from which interaction-related audio segments are filtered using the GPT-4o mini. 
The output for these ignored instances is selected from a set of ten predefined sentences provided in Appendix~\ref{sec:appendix_4}.

During training, the loss of the state tokens is included.
The ConMamba encoder is frozen, and only the Mamba language model and speech adapter are trained.
Assuming the index of the state token in the prompt is $\bm{j}$, the loss function is calculated as follows.
\begin{align}
\mathcal{L}_1 &= - \log P(\mathcal{T}(\bm{S})_{j} \mid \mathcal{T}(\bm{S})_{<j}) \label{eq:L1} \\
\mathcal{L}_2 &= - \sum_{i=1}^{M} \log P(\bm{y}_i^T \mid \mathcal{T}(\bm{S}), \bm{Y}_{<i}^T)  \label{eq:L2} \\
\mathcal{L} \hspace{5pt} &= \mathcal{L}_1 +\mathcal{L}_2 
\label{eq:L}
\end{align}

\paragraph{Streaming Alignment}
\label{streaming_alignment}
The ConMamba encoder requires bidirectional feature computation, which inherently requires offline audio encoding.
To address this limitation and enable real-time interaction, our model is specifically designed to process speech slices dynamically, without waiting for the complete speech input. 
Through innovative streaming alignment techniques~\cite{yao2024minicpm}, we ensure real-time processing while preserving the comprehensive feature computation capabilities of the bidirectional ConMamba encoder.

The slice size critically influences our model's performance. 
Following the prior research~\cite{zhang2024beyond}, we slice the speech data used during the input state discrimination stage into 3-second intervals, feeding these precise slices into the model for refined fine-tuning.

The loss functions are computed as illustrated in Equations (\ref{eq:L1}, \ref{eq:L2}, \ref{eq:L}). 
Notably, we maintain the Mamba language model in a frozen state, focusing our training exclusively on the ConMamba encoder and speech adapter.

\begin{figure*}[t!]
    \centering
    \includegraphics[width=\linewidth]{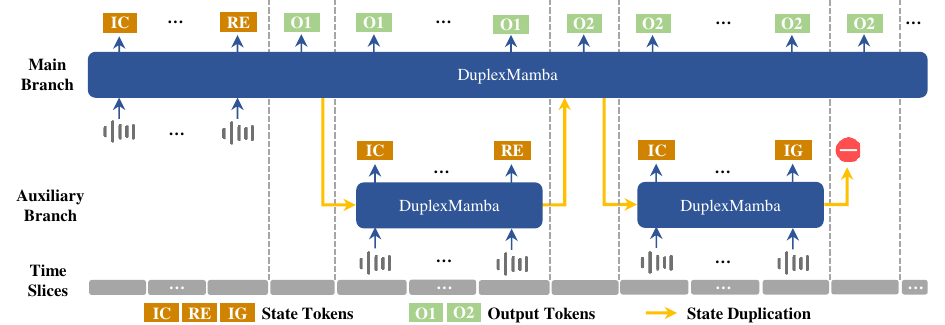}
    \caption{The duplex decoding strategy of DuplexMamba. "IC" is short for the "<incomplete>" token, "RE" for the "<response>" token, and "IG" for the "<ignore>" token. "O1" and "O2" represent the output tokens for query 1 and query 2, respectively. Due to the fixed state size in Mamba-based models, creating an auxiliary branch simply involves duplicating the model's current state.}
    \label{fig:inference}
\end{figure*}

\subsection{Duplex Decoding}
To enhance user experience, \citet{fu2024vita} introduces two interaction paradigms: \textbf{non-awakening interaction} and \textbf{interruption interaction}, which demand duplex decoding capability. 
In non-awakening interaction, the model is designed to prevent responding to background dialogues or non-query inputs. 
Conversely, interruption interaction enables the model to suspend its ongoing output and immediately process and respond to the most recent user input when a query is detected.

Our duplex decoding strategy is illustrated in Figure~\ref{fig:inference}.
To enable real-time processing capability, both input and output streams are structured in slice format. 
Following \citet{zhang2024beyond}, we implement time-sliced chunking at 2-3 second intervals, with each slice containing approximately 4-6 words.
During each processing interval, the model predicts the state token of the current input or generates a response segment.
If the model receives a new input while generating a response segment, the model's state is duplicated to create an auxiliary decoding branch.
The auxiliary decoding branch processes the new input and predicts the new input's state.

\paragraph{Decoding Branch Creation}
When a user submits a new speech input during the model's ongoing generation process, the system creates an auxiliary decoding branch by duplicating the state of the Mamba-based language model.
The system then concatenates the "<eos> <$|$user$|$>" tokens with the new input, enabling the auxiliary decoding branch to process it while the main branch continues to generate the response concurrently.
Due to the fixed state size in Mamba-based models, creating an auxiliary decoding branch simply requires duplicating the model's existing state.

\paragraph{Decoding Branch Switching}
\label{branch_switching}
As described in Section~\ref{state_discrimination}, the model is capable of predicting the input state.
If the auxiliary branch generates an "<incomplete>" token, the model's state is rolled back to await the next input slice. 
When the auxiliary branch generates a "<response>" token, it transitions to the main branch while overwriting the main branch model's state.
The main branch then processes the "<$|$assistant$|$>" tokens, allowing it to smoothly transition into generating response segments for the new query in subsequent time slices. 
Conversely, if the auxiliary branch generates an "<ignore>" token, the new input requires no response, and the auxiliary branch is discarded.

\section{Experiments}

\subsection{Setup}
\label{sec6}
\paragraph{Training Data}
The training data for the four stages is summarized in Table~\ref{tab:training_data}.

In stage 1, the model is trained on ASR data from LibriSpeech \cite{panayotov2015librispeech}, TED-LIUM 3 \cite{hernandez2018ted}, and Multilingual LibriSpeech~\cite{pratap2020mls}, totaling around 11k hours. 
All data is normalized to lowercase English text without punctuation.

In stage 2, the training tasks include ASR and speech-to-text QA tasks. 
The ASR data contains 50k samples from Multilingual LibriSpeech.
The QA dataset, VoiceAssistant~\cite{xie2024mini}, is used for fine-tuning speech models in Mini-Omni. 
Following \citet{fang2024llama}, we remove identity-related data and retain only the first-round instruction from multi-turn conversations, resulting in approximately 200k training samples.

\begin{table}[t!]
\centering
\scalebox{0.85}{
\begin{tabular}{cclc}
\toprule
\textbf{Stage}       & \textbf{Task}        & \textbf{Dataset}         & \multicolumn{1}{c}{\textbf{Items}} \\ 
\midrule
\multirow{3}{*}{1}   & \multirow{3}{*}{ASR} & LibriSpeech              & 960h                      \\
                     &                      & TED-LIUM 3               & 450h                      \\
                     &                      & Multilingual LibriSpeech & 10000h                    \\ 
\midrule
\multirow{2}{*}{2}   & ASR                  & Multilingual LibriSpeech & 50k                       \\
                     & QA                   & VoiceAssistant           & 200k                      \\ 
\midrule
\multirow{4}{*}{3,4} & ASR                  & Multilingual LibriSpeech & 5k                        \\ 
\cline{2-4}
                     & \multirow{3}{*}{QA}  & Response-Required Data         & 10k                       \\
                     &                      & Incomplete Data             & 5k                        \\
                     &                      & Ingored Data              & 5k                        \\ 
\bottomrule
\end{tabular}}
\caption{Training data for each stage.}
\label{tab:training_data}
\end{table}

\begin{table*}[t!]
\centering
\scalebox{0.78}{
\begin{tabular}{lcccccccc}
\toprule
\multirow{2}{*}{\textbf{Model}}  & \textbf{AlpacaEval} & \textbf{CommonEval} & \multicolumn{2}{c}{\textbf{SD-QA}} & \multicolumn{2}{c}{\textbf{IFEval}}      & \textbf{AdvBench}       & \multirow{2}{*}{\textbf{Overall}} \\
                  & \textbf{(GPT)}  & \textbf{(GPT)}  & \textbf{(Panda)}    & \textbf{(GPT)}    & \textbf{(P. Acc.)} & \textbf{(I. Acc.)} & \textbf{(Refusal Rate)} &                          \\ 
\midrule
\multicolumn{9}{c}{non-duplex models}                                \\
\midrule
DiVA              & 3.67       & \textbf{3.54}   & \textbf{62.39}  & \textbf{51.72}     & \textbf{34.93}    & \textbf{43.38}    & \textbf{98.27}     & \textbf{67.73}                    \\
LLaMA-Omni        & 3.70       & 3.46       & 40.14      & 39.24        & 10.15         & 19.58           & 11.35          & 41.83                    \\
Mini-Omni         & 1.95       & 2.02       & 23.69      & 4.16         & 8.99          & 18.17           & 37.12          & 28.80                    \\
Qwen2-Audio       & \textbf{3.74}   & 3.43       & 41.77      & 29.66        & 20.73         & 31.93           & 96.73          & 60.45                    \\
\midrule
\multicolumn{9}{c}{duplex models}                                \\
\midrule
VITA             & \textbf{3.38}   & 2.15       & \textbf{31.28}  & \textbf{24.59}    & \textbf{18.12}    & \textbf{27.51}     & 26.73          & 37.62                    \\
Moshi            & 2.01       & 1.60       & 15.01      & 16.27        & 6.38          & 13.76           & 44.23          & 28.43                    \\
Mini-Omni2        & 2.32       & 2.18       & 11.03      & 7.59         & 7.25          & 15.86           & 57.50          & 33.67                    \\
DuplexMamba~(ours) & 3.18       & \textbf{2.95}   & 17.18      & 19.35        & 12.46         & 20.68           & \textbf{93.85}   & \textbf{50.26}             \\ 
\bottomrule
\end{tabular}}
\caption{The performance of various voice assistant models on VoiceBench. Higher values are better for all metrics.}
\label{tab:voicebench}
\end{table*}
 
In stages 3 and 4, the training data includes ASR and the state discrimination dataset. 
The ASR data consists of 5k samples from the Multilingual LibriSpeech.
As described in Section~\ref{state_discrimination}, the state discrimination dataset is comprised of three types of data: response-required, incomplete, and ignored.
The response-required data and incomplete data are extracted from the VoiceAssistant dataset. 
The ignored data is sourced from the podcast, the audiobook, and the YouTube content in the GigaSpeech L~\cite{chen2021gigaspeech} dataset. 

The training details are listed in Section~\ref{sec.hyper}.

\paragraph{Baselines}
We compare the performance of our model with various end-to-end voice assistant models, including Qwen2-Audio~\cite{chu2024qwen2}, LLaMA-Omni~\cite{fang2024llama}, Mini-Omni~\cite{xie2024mini}, Mini-Omni2~\cite{xie2024mini2}, VITA~\cite{fu2024vita}, Moshi~\cite{defossez2024moshi}, and DiVA~\cite{held2024distilling}.
The architectures of these models are summarized in Table~\ref{tab:model_architecture}.

\paragraph{Evaluation}
We evaluate DuplexMamba's performance on VoiceBench~\cite{chen2024voicebench}, a benchmark designed to assess LLM-based voice assistant models.
VoiceBench focuses on real-world challenges, including diverse speakers, varying environmental conditions, and different content types.
Evaluation metrics for all models focus on the quality of text responses. 

VoiceBench includes several evaluation tasks.
For open-ended question-answering tasks (AlpacaEval and CommonEval), responses are scored from 1 to 5 by GPT based on ground truth instructions. 
In SD-QA, accuracy is evaluated using human-labeled reference answers and assessed via both PANDA and GPT methods.
For IFEval, we follow ~\citet{zhou2023instruction} to compute both loose and strict accuracy, reporting their average at both the prompt and instruction levels.
AdvBench measures safety based on the refusal rate, where higher rates indicate safer models.
All GPT-based evaluations are conducted using GPT-4o mini.

\subsection{Main Results}
\label{sec4-2}
Table~\ref{tab:voicebench} presents the performance of DuplexMamba and other end-to-end voice assistant models on the VoiceBench datasets.
Overall, DuplexMamba achieves the best performance among duplex models, demonstrating the effectiveness of our proposed model.

Moreover, our model ranks third among all the models, behind DiVA and Qwen2-Audio, which are Transformer-based models with 8B and 7B parameters, respectively.
Despite having only 2.8B parameters, our Mamba-based model, DuplexMamba, outperforms voice assistant models like VITA and LLaMA-Omni, both of which rely on Transformer architectures with 7B and 8B parameters. 
This highlights the impressive potential of the Mamba architecture.

The performance of DuplexMamba on the SD-QA task is relatively poor. 
VoiceBench selects a subset of oral questions from the original dataset, removes contextual information, and requires voice assistant models to respond using internal knowledge. 
Since our fine-tuning data provides limited internal knowledge, performance on this task primarily depends on pre-training. 
We anticipate that using a larger-scale language model with more pre-training data will significantly improve the results.

\subsection{Analysis}
\label{sec4-3}

\paragraph{ASR Performance}
We evaluate our model's performance on ASR tasks.
Experiments are conducted using two test sets and two validation sets from LibriSpeech: test-clean, test-other, dev-clean, and dev-other.
The results are presented in Table~\ref{table.asr}.
"Stage 2" refers to our model trained through two stages, representing a voice assistant without duplex and streaming encoding capabilities.
"Stage 4" refers to our model after four stages of training, which incorporates both streaming encoding and duplex capabilities.
The compared models include non-duplex models such as wav2vec2~\cite{baevski2020wav2vec}, Whisper-small~\cite{radford2023robust}, and Mini-Omni, as well as the duplex model VITA.

The results indicate that among all non-duplex models, the ASR performance of our DuplexMamba at stage 2 is only slightly behind that of the Whisper-small decoder while still demonstrating strong speech comprehension.
Although training in stages 3 and 4 leads to a slight reduction in ASR performance, the DuplexMamba at stage 4 remains highly competitive on ASR benchmarks, outperforming VITA, another duplex model.

\begin{table}[t!]
\centering
\scalebox{0.71}{
\begin{tabular}{lcccc}
\toprule
\textbf{Model} & \textbf{test-clean} & \textbf{test-other} & \textbf{dev-clean} & \textbf{dev-other} \\ 
\midrule
wav2vec2-base  & 6.00                & 13.40               & -                  & -                  \\
VITA          & 8.14                & 18.41               & 7.57               & 16.57              \\
Whisper-small  & 3.40                & 7.60                & -                  & -                  \\
Mini-Omni      & 4.50                & 9.70                & 4.60               & 9.20               \\
DuplexMamba   &                &               &              &              \\
\;\;\;\;stage 2   & 3.36                & 8.23                & 3.38               & 8.26               \\
\;\;\;\;stage 4  & 4.94                & 10.82               & 4.90               & 10.83              \\ 
\bottomrule
\end{tabular}}
\caption{Comparison of DuplexMamba's ASR performance with other models. VITA and DuplexMamba at stage 4 are duplex models.}
\label{table.asr}
\end{table}

\paragraph{Decoding Efficiency}
Due to Mamba's linear complexity, DuplexMamba operates as a streaming model.
To assess its effectiveness, we compare the GPU memory usage of Qwen2-Audio and DuplexMamba across different context lengths.

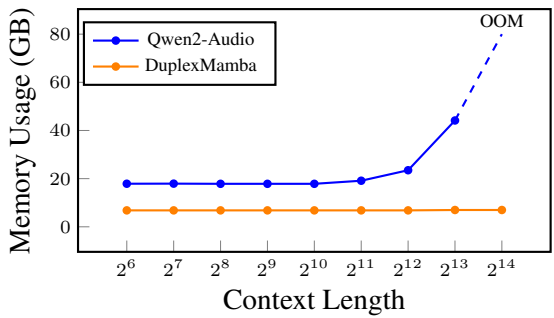
\begin{figure}[ht!]
   \centering
    \pgfplotsset{height=4.8cm,width=7.8cm,compat=1.14,every axis/.append style={thick},legend columns=1}    
        \begin{tikzpicture}
        \tikzset{every node}=[font=\scriptsize]
        \begin{axis}
		[enlargelimits=0.13, tick align=inside, 
        xtick pos=left, %
        ytick pos=left, %
        xticklabels={$2^6$, $2^7$, $2^8$, $2^9$, $2^{10}$, $2^{11}$, $2^{12}$, $2^{13}$, $2^{14}$},
		xtick={6, 7, 8, 9, 10, 11, 12, 13, 14},
		ymin=0,
		ymax=80,
		legend style={at={(0.42,0.95)},
        legend columns=1,
        },
		ylabel={\normalsize Memory Usage (GB)},xlabel={\normalsize Context Length },font=\scriptsize]

		\addplot+ [sharp plot,mark=*,mark size=1.2pt,mark options={solid,mark color=black}, color=blue] coordinates
            {(6,17.888)(7,17.924)(8,17.852)(9,17.852)(10,17.852)(11,19.134)(12,23.490)(13,44.148)};
		\addlegendentry{\scriptsize Qwen2-Audio\;}

        \addplot+ [sharp plot,mark=*,mark size=1.2pt,mark options={solid,mark color=orange}, color=orange] coordinates
            {(6,6.824)(7,6.824)(8,6.824)(9,6.824)(10,6.824)(11,6.824)(12,6.824)(13,6.978)(14,6.978)};
		\addlegendentry{\scriptsize DuplexMamba\;}

            \addplot+ [sharp plot,mark=none,mark size=1.2pt,mark options={solid,mark color=black}, color=blue, style=dashed] coordinates
            {(13,44.148)(14,80)};

            \node at (axis cs:14,80) [anchor=south] {\scriptsize OOM};

		\end{axis}
        \end{tikzpicture}
	\caption{GPU memory usage of DuplexMamba and Qwen2-Audio across different context lengths. }
	\label{fig.memory}
\end{figure}

All experiments are conducted on a single NVIDIA A100 GPU with 80GB of memory. 
Each test is repeated five times, and we report the average results. The results are presented in Figure~\ref{fig.memory}.
As the context length increases, the memory usage of the Transformer-based model, Qwen2-Audio, grows rapidly. When the context length reaches 16384, the GPU runs out of memory. In contrast, our Mamba-based model maintains stable memory usage due to its fixed-size state, demonstrating superior memory efficiency.

GPU memory usage directly affects decoding efficiency, with optimization improving speed and enabling longer context processing.

\paragraph{Interruption and Non-awakening}
The performance of interruption-ignoring is critical to the quality of duplex interactions. 
We apply the method described in Section~\ref{state_discrimination} to generate 1.7k response-required samples and 500 ignore-needed samples for testing. 
Duplex models, VITA and DuplexMamba, generate state tokens to distinguish between interruption and non-awakening interaction, while non-duplex model Qwen2-Audio directly generates "interrupt" or "ignore" using a zero-shot approach.

We classify the model's generation of state tokens or text indicating "interrupt" as the positive class and "ignore" as the negative class. 
We then compute precision, recall, and F1 score.

The experimental results, presented in Table~\ref{tab:interruption_ignoring}, show that DuplexMamba outperforms Qwen2-Audio and VITA in all metrics. 
This highlights the effectiveness of our input state discrimination training stage and establishes a critical foundation for our duplex decoding strategy.

\begin{table}[t!]
\centering
\scalebox{0.89}{
\begin{tabular}{lccc}
\toprule
                  & Precision & Recall & F1      \\ \midrule
Qwen2-Audio       & 74.52     & 84.79  & 79.32   \\
VITA              & 93.36     & 87.76  & 90.47   \\
DuplexMamba(ours) & 99.46     & 99.23  & 99.35   \\ \bottomrule
\end{tabular}}
\caption{Comparison of DuplexMamba's Interruption and Non-awakening performance with other models.}
\label{tab:interruption_ignoring}
\end{table}

\subsection{Case Study}
\begin{figure*}[t!]
    \centering
    \includegraphics[width=\linewidth]{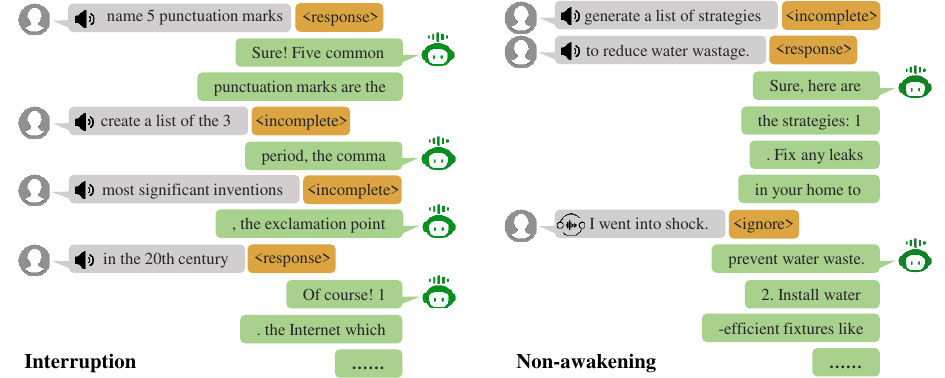}
    \caption{Cases of interruption interaction and non-awakening interaction. The model predicts the state token for each user input.}
    \label{fig:case_study}
\end{figure*}

In Figure~\ref{fig:case_study}, we illustrate cases of the two interaction paradigms.
Our model processes inputs and generates outputs in a time-slice manner.
In the case of interruption interaction, when the user inputs "create a list of the 3", the model continues the ongoing generation while simultaneously processing the new input and predicting whether and when to respond.
When the user then inputs "in the 20th century", the model predicts a "<response>" token and seamlessly transitions to respond to the new input.
In the case of non-awakening interaction, the model automatically filters out the background dialogue "I went into shock." by predicting an "<ignore>" token.

\section{Related Work}
\label{sec5}
\subsection{Real-Time Speech Interaction Models}
Real-time speech interaction models can be classified into non-duplex models and duplex models.

\paragraph{Non-Duplex Models}
There are two architectures: cascaded models and end-to-end models.

For cascaded models, HuggingGPT~\cite{shen2024hugginggpt} facilitates task decomposition of human instructions by LLMs and invokes models from Huggingface to perform specific tasks, including various ASR models. 
Audiogpt~\cite{huang2024audiogpt} leverages multiple audio models to process complex audio information, linking the LLM with an input interface (ASR) for speech interactions.

For end-to-end models, SpeechGPT~\cite{zhang2023speechgpt} and AudioPaLM~\cite{rubenstein2023audiopalm} integrate speech tokens into the LLM’s vocabulary, continuing pretraining with both speech and text data. 
Qwen2-Audio~\cite{chu2024qwen2} and SALMONN~\cite{tangsalmonn} involve adding a speech encoder before the LLM and conducting multi-stage training. 
LLaMA-Omni~\cite{fang2024llama} and Mini-Omni~\cite{xie2024mini} further incorporate speech adapters between the speech encoder and LLM. 
DiVA~\cite{held2024distilling} trains speech-based LLMs without instruction data by using a text LLM’s responses to transcribed text for self-supervised cross-modal distillation.

\paragraph{Duplex Models}
These models can process new user inputs while generating responses simultaneously~\cite{veluri2024beyond,xu2024enabling}.

MiniCPM-duplex~\cite{zhang2024beyond} uses time-division multiplexing to process queries and responses in time slices for pseudo-simultaneous interaction.
LSLM~\cite{ma2024language} detects real-time turn-taking by combining input and output tokens for autoregressive generation.
Moshi~\cite{defossez2024moshi} enables parallel processing by modeling both input and output speech streams concurrently.
SyncLLM~\cite{veluri2024beyond} processes tokens from both streams concurrently using an interleaved approach.
Freeze-Omni~\cite{wang2024freeze} supports low-latency speech-to-speech interaction with a frozen backbone LLM to prevent catastrophic forgetting.
VITA~\cite{fu2024vita} alternates between two models for duplex interaction, using state tokens to distinguish effective from non-effective queries.

\subsection{Streaming Architectures}
Streaming architectures have linear complexity with input length and can generally be categorized into Linear RNN and Linear Attention models.

Linear RNN models like Mamba~\cite{gu2023mamba} and Mamba-2~\cite{daotransformers} optimize RNNs for specific hardware, enabling efficient training.
During inference, they process sequences step-by-step, maintaining a fixed-size context state, which ensures high memory efficiency and low latency for long-context tasks.

Linear Attention models, such as RWKV~\cite{peng2023rwkv} and RetNet~\cite{sun2023retentive}, eliminate certain nonlinear dependencies in the Attention mechanism, making them as efficient as RNNs during inference.

\section{Conclusion}
\label{secc}
We propose DuplexMamba, an end-to-end multimodal duplex model designed for real-time speech-to-text conversation.
DuplexMamba integrates a Mamba-based speech encoder with a Mamba-based language model, enabling simultaneous input processing and output generation through a novel duplex decoding strategy.
To achieve duplex and streaming capabilities, we incorporate input state discrimination and streaming alignment training stages.
Experiments demonstrate that DuplexMamba achieves performance comparable to several recently developed Transformer-based models in ASR and voice assistant tasks.

\section*{Limitations}

The pre-trained Mamba-based language model used in our model has 2.8B parameters, which is smaller than the 7B parameters commonly used in mainstream LLMs, limiting its foundational capabilities. 
Additionally, we utilize the first version of the Mamba architecture; replacing it with the Mamba-2 architecture could further enhance the model's processing speed.

\section*{Ethical Statement}

Our research relies on publicly accessible models and well-documented datasets, all of which are properly cited. 
By utilizing widely recognized datasets with established safety standards, we effectively reduce the risk of generating harmful content.

\bibliography{custom}

\begin{thebibliography}{40}
\providecommand{\natexlab}[1]{#1}

\bibitem[{Achiam et~al.(2023)Achiam, Adler, Agarwal, Ahmad, Akkaya, Aleman, Almeida, Altenschmidt, Altman, Anadkat et~al.}]{achiam2023gpt}
Josh Achiam, Steven Adler, Sandhini Agarwal, Lama Ahmad, Ilge Akkaya, Florencia~Leoni Aleman, Diogo Almeida, Janko Altenschmidt, Sam Altman, Shyamal Anadkat, et~al. 2023.
\newblock Gpt-4 technical report.
\newblock \emph{arXiv preprint arXiv:2303.08774}.

\bibitem[{Baevski et~al.(2020)Baevski, Zhou, Mohamed, and Auli}]{baevski2020wav2vec}
Alexei Baevski, Yuhao Zhou, Abdelrahman Mohamed, and Michael Auli. 2020.
\newblock wav2vec 2.0: A framework for self-supervised learning of speech representations.
\newblock \emph{Advances in neural information processing systems}, pages 12449--12460.

\bibitem[{Chen et~al.(2021)Chen, Chai, Wang, Du, Zhang, Weng, Su, Povey, Trmal, Zhang et~al.}]{chen2021gigaspeech}
Guoguo Chen, Shuzhou Chai, Guanbo Wang, Jiayu Du, Wei~Qiang Zhang, Chao Weng, Dan Su, Daniel Povey, Jan Trmal, Junbo Zhang, et~al. 2021.
\newblock Gigaspeech: An evolving, multi-domain asr corpus with 10,000 hours of transcribed audio.
\newblock In \emph{Annual Conference of the International Speech Communication Association}, pages 4376--4380.

\bibitem[{Chen et~al.(2024)Chen, Yue, Zhang, Gao, Tan, and Li}]{chen2024voicebench}
Yiming Chen, Xianghu Yue, Chen Zhang, Xiaoxue Gao, Robby~T Tan, and Haizhou Li. 2024.
\newblock Voicebench: Benchmarking llm-based voice assistants.
\newblock \emph{arXiv preprint arXiv:2410.17196}.

\bibitem[{Chu et~al.(2024)Chu, Xu, Yang, Wei, Wei, Guo, Leng, Lv, He, Lin et~al.}]{chu2024qwen2}
Yunfei Chu, Jin Xu, Qian Yang, Haojie Wei, Xipin Wei, Zhifang Guo, Yichong Leng, Yuanjun Lv, Jinzheng He, Junyang Lin, et~al. 2024.
\newblock Qwen2-audio technical report.
\newblock \emph{arXiv preprint arXiv:2407.10759}.

\bibitem[{Dao and Gu(2024)}]{daotransformers}
Tri Dao and Albert Gu. 2024.
\newblock Transformers are ssms: Generalized models and efficient algorithms through structured state space duality.
\newblock In \emph{International Conference on Machine Learning}.

\bibitem[{D{\'e}fossez et~al.(2024)D{\'e}fossez, Mazar{\'e}, Orsini, Royer, P{\'e}rez, J{\'e}gou, Grave, and Zeghidour}]{defossez2024moshi}
Alexandre D{\'e}fossez, Laurent Mazar{\'e}, Manu Orsini, Am{\'e}lie Royer, Patrick P{\'e}rez, Herv{\'e} J{\'e}gou, Edouard Grave, and Neil Zeghidour. 2024.
\newblock Moshi: a speech-text foundation model for real-time dialogue.
\newblock \emph{arXiv preprint arXiv:2410.00037}.

\bibitem[{Fang et~al.(2024)Fang, Guo, Zhou, Ma, Zhang, and Feng}]{fang2024llama}
Qingkai Fang, Shoutao Guo, Yan Zhou, Zhengrui Ma, Shaolei Zhang, and Yang Feng. 2024.
\newblock Llama-omni: Seamless speech interaction with large language models.
\newblock \emph{arXiv preprint arXiv:2409.06666}.

\bibitem[{Fu et~al.(2024)Fu, Lin, Long, Shen, Zhao, Zhang, Dong, Wang, Yin, Ma et~al.}]{fu2024vita}
Chaoyou Fu, Haojia Lin, Zuwei Long, Yunhang Shen, Meng Zhao, Yifan Zhang, Shaoqi Dong, Xiong Wang, Di~Yin, Long Ma, et~al. 2024.
\newblock Vita: Towards open-source interactive omni multimodal llm.
\newblock \emph{arXiv preprint arXiv:2408.05211}.

\bibitem[{Gu and Dao(2023)}]{gu2023mamba}
Albert Gu and Tri Dao. 2023.
\newblock Mamba: Linear-time sequence modeling with selective state spaces.
\newblock \emph{arXiv preprint arXiv:2312.00752}.

\bibitem[{Held et~al.(2024)Held, Li, Ryan, Shi, Zhang, and Yang}]{held2024distilling}
William Held, Ella Li, Michael Ryan, Weiyan Shi, Yanzhe Zhang, and Diyi Yang. 2024.
\newblock Distilling an end-to-end voice assistant without instruction training data.
\newblock \emph{arXiv preprint arXiv:2410.02678}.

\bibitem[{Hernandez et~al.(2018)Hernandez, Nguyen, Ghannay, Tomashenko, and Esteve}]{hernandez2018ted}
Fran{\c{c}}ois Hernandez, Vincent Nguyen, Sahar Ghannay, Natalia Tomashenko, and Yannick Esteve. 2018.
\newblock Ted-lium 3: Twice as much data and corpus repartition for experiments on speaker adaptation.
\newblock In \emph{Speech and Computer: International Conference}, pages 198--208.

\bibitem[{Huang et~al.(2024)Huang, Li, Yang, Shi, Chang, Ye, Wu, Hong, Huang, Liu et~al.}]{huang2024audiogpt}
Rongjie Huang, Mingze Li, Dongchao Yang, Jiatong Shi, Xuankai Chang, Zhenhui Ye, Yuning Wu, Zhiqing Hong, Jiawei Huang, Jinglin Liu, et~al. 2024.
\newblock Audiogpt: Understanding and generating speech, music, sound, and talking head.
\newblock In \emph{Proceedings of the AAAI Conference on Artificial Intelligence}, pages 23802--23804.

\bibitem[{Jiang et~al.(2024)Jiang, Li, Florea, Han, and Mesgarani}]{jiang2024speech}
Xilin Jiang, Yinghao~Aaron Li, Adrian~Nicolas Florea, Cong Han, and Nima Mesgarani. 2024.
\newblock Speech slytherin: Examining the performance and efficiency of mamba for speech separation, recognition, and synthesis.
\newblock \emph{arXiv preprint arXiv:2407.09732}.

\bibitem[{LeCun et~al.(1998)LeCun, Bottou, Bengio, and Haffner}]{lecun1998gradient}
Yann LeCun, L{\'e}on Bottou, Yoshua Bengio, and Patrick Haffner. 1998.
\newblock Gradient-based learning applied to document recognition.
\newblock \emph{Proceedings of the IEEE}, pages 2278--2324.

\bibitem[{Ma et~al.(2024{\natexlab{a}})Ma, Song, Du, Cong, Chen, Wang, Wang, and Chen}]{ma2024language}
Ziyang Ma, Yakun Song, Chenpeng Du, Jian Cong, Zhuo Chen, Yuping Wang, Yuxuan Wang, and Xie Chen. 2024{\natexlab{a}}.
\newblock Language model can listen while speaking.
\newblock \emph{arXiv preprint arXiv:2408.02622}.

\bibitem[{Ma et~al.(2024{\natexlab{b}})Ma, Yang, Yang, Gao, Wang, Du, Yu, Chen, Zheng, Zhang et~al.}]{ma2024embarrassingly}
Ziyang Ma, Guanrou Yang, Yifan Yang, Zhifu Gao, Jiaming Wang, Zhihao Du, Fan Yu, Qian Chen, Siqi Zheng, Shiliang Zhang, et~al. 2024{\natexlab{b}}.
\newblock An embarrassingly simple approach for llm with strong asr capacity.
\newblock \emph{arXiv preprint arXiv:2402.08846}.

\bibitem[{OpenAI(2022)}]{2022chatgpt}
OpenAI. 2022.
\newblock Introducing chatgpt.
\newblock \url{https://openai.com/blog/chatgpt}.

\bibitem[{OpenAI(2024)}]{2024gpt4o}
OpenAI. 2024.
\newblock Hello gpt-4o.
\newblock \url{https://openai.com/index/hello-gpt-4o/}.

\bibitem[{Panayotov et~al.(2015)Panayotov, Chen, Povey, and Khudanpur}]{panayotov2015librispeech}
Vassil Panayotov, Guoguo Chen, Daniel Povey, and Sanjeev Khudanpur. 2015.
\newblock Librispeech: an asr corpus based on public domain audio books.
\newblock In \emph{IEEE international conference on acoustics, speech and signal processing}, pages 5206--5210.

\bibitem[{Peng et~al.(2023)Peng, Alcaide, Anthony, Albalak, Arcadinho, Biderman, Cao, Cheng, Chung, Derczynski et~al.}]{peng2023rwkv}
Bo~Peng, Eric Alcaide, Quentin~Gregory Anthony, Alon Albalak, Samuel Arcadinho, Stella Biderman, Huanqi Cao, Xin Cheng, Michael~Nguyen Chung, Leon Derczynski, et~al. 2023.
\newblock Rwkv: Reinventing rnns for the transformer era.
\newblock In \emph{The Conference on Empirical Methods in Natural Language Processing}.

\bibitem[{Pratap et~al.(2020)Pratap, Xu, Sriram, Synnaeve, and Collobert}]{pratap2020mls}
Vineel Pratap, Qiantong Xu, Anuroop Sriram, Gabriel Synnaeve, and Ronan Collobert. 2020.
\newblock Mls: A large-scale multilingual dataset for speech research.
\newblock \emph{Interspeech}.

\bibitem[{Qian et~al.(2024)Qian, Liu, Liu, Chen, Dang, Li, Yang, Chen, Su, Cong et~al.}]{qian2024chatdev}
Chen Qian, Wei Liu, Hongzhang Liu, Nuo Chen, Yufan Dang, Jiahao Li, Cheng Yang, Weize Chen, Yusheng Su, Xin Cong, et~al. 2024.
\newblock Chatdev: Communicative agents for software development.
\newblock In \emph{Proceedings of the Annual Meeting of the Association for Computational Linguistics}, pages 15174--15186.

\bibitem[{Radford et~al.(2023)Radford, Kim, Xu, Brockman, McLeavey, and Sutskever}]{radford2023robust}
Alec Radford, Jong~Wook Kim, Tao Xu, Greg Brockman, Christine McLeavey, and Ilya Sutskever. 2023.
\newblock Robust speech recognition via large-scale weak supervision.
\newblock In \emph{International conference on machine learning}, pages 28492--28518.

\bibitem[{Rubenstein et~al.(2023)Rubenstein, Asawaroengchai, Nguyen, Bapna, Borsos, Quitry, Chen, Badawy, Han, Kharitonov et~al.}]{rubenstein2023audiopalm}
Paul~K Rubenstein, Chulayuth Asawaroengchai, Duc~Dung Nguyen, Ankur Bapna, Zal{\'a}n Borsos, F{\'e}lix de~Chaumont Quitry, Peter Chen, Dalia~El Badawy, Wei Han, Eugene Kharitonov, et~al. 2023.
\newblock Audiopalm: A large language model that can speak and listen.
\newblock \emph{arXiv preprint arXiv:2306.12925}.

\bibitem[{Shen et~al.(2024)Shen, Song, Tan, Li, Lu, and Zhuang}]{shen2024hugginggpt}
Yongliang Shen, Kaitao Song, Xu~Tan, Dongsheng Li, Weiming Lu, and Yueting Zhuang. 2024.
\newblock Hugginggpt: Solving ai tasks with chatgpt and its friends in hugging face.
\newblock \emph{Advances in Neural Information Processing Systems}.

\bibitem[{Sun et~al.(2023)Sun, Dong, Huang, Ma, Xia, Xue, Wang, and Wei}]{sun2023retentive}
Yutao Sun, Li~Dong, Shaohan Huang, Shuming Ma, Yuqing Xia, Jilong Xue, Jianyong Wang, and Furu Wei. 2023.
\newblock Retentive network: A successor to transformer for large language models.
\newblock \emph{arXiv preprint arXiv:2307.08621}.

\bibitem[{Tang et~al.(2024)Tang, Yu, Sun, Chen, Tan, Li, Lu, Zejun, and Zhang}]{tangsalmonn}
Changli Tang, Wenyi Yu, Guangzhi Sun, Xianzhao Chen, Tian Tan, Wei Li, Lu~Lu, MA~Zejun, and Chao Zhang. 2024.
\newblock Salmonn: Towards generic hearing abilities for large language models.
\newblock In \emph{International Conference on Learning Representations}.

\bibitem[{Vaswani(2017)}]{vaswani2017attention}
A~Vaswani. 2017.
\newblock Attention is all you need.
\newblock \emph{Advances in Neural Information Processing Systems}.

\bibitem[{Veluri et~al.(2024)Veluri, Peloquin, Yu, Gong, and Gollakota}]{veluri2024beyond}
Bandhav Veluri, Benjamin Peloquin, Bokai Yu, Hongyu Gong, and Shyamnath Gollakota. 2024.
\newblock Beyond turn-based interfaces: Synchronous llms as full-duplex dialogue agents.
\newblock In \emph{Proceedings of the Conference on Empirical Methods in Natural Language Processing}, pages 21390--21402.

\bibitem[{Wang et~al.(2023b)Wang, Xie, Jiang, Mandlekar, Xiao, Zhu, Fan, and Anandkumar}]{wangvoyager}
Guanzhi Wang, Yuqi Xie, Yunfan Jiang, Ajay Mandlekar, Chaowei Xiao, Yuke Zhu, Linxi Fan, and Anima Anandkumar. 2023b.
\newblock Voyager: An open-ended embodied agent with large language models.
\newblock In \emph{Intrinsically-Motivated and Open-Ended Learning Workshop on Neural Information Processing Systems}.

\bibitem[{Wang et~al.(2024)Wang, Li, Fu, Shen, Xie, Li, Sun, and Ma}]{wang2024freeze}
Xiong Wang, Yangze Li, Chaoyou Fu, Yunhang Shen, Lei Xie, Ke~Li, Xing Sun, and Long Ma. 2024.
\newblock Freeze-omni: A smart and low latency speech-to-speech dialogue model with frozen llm.
\newblock \emph{arXiv preprint arXiv:2411.00774}.

\bibitem[{Xie and Wu(2024{\natexlab{a}})}]{xie2024mini}
Zhifei Xie and Changqiao Wu. 2024{\natexlab{a}}.
\newblock Mini-omni: Language models can hear, talk while thinking in streaming.
\newblock \emph{arXiv preprint arXiv:2408.16725}.

\bibitem[{Xie and Wu(2024{\natexlab{b}})}]{xie2024mini2}
Zhifei Xie and Changqiao Wu. 2024{\natexlab{b}}.
\newblock Mini-omni2: Towards open-source gpt-4o with vision, speech and duplex capabilities.
\newblock \emph{arXiv preprint arXiv:2410.11190}.

\bibitem[{Xu et~al.(2024)Xu, Wang, Zhao, Han, Yan, Zhang, Tao, Liu, and Che}]{xu2024enabling}
Wang Xu, Shuo Wang, Weilin Zhao, Xu~Han, Yukun Yan, Yudi Zhang, Zhe Tao, Zhiyuan Liu, and Wanxiang Che. 2024.
\newblock Enabling real-time conversations with minimal training costs.
\newblock \emph{arXiv preprint arXiv:2409.11727}.

\bibitem[{Yao et~al.(2024)Yao, Yu, Zhang, Wang, Cui, Zhu, Cai, Li, Zhao, He et~al.}]{yao2024minicpm}
Yuan Yao, Tianyu Yu, Ao~Zhang, Chongyi Wang, Junbo Cui, Hongji Zhu, Tianchi Cai, Haoyu Li, Weilin Zhao, Zhihui He, et~al. 2024.
\newblock Minicpm-v: A gpt-4v level mllm on your phone.
\newblock \emph{arXiv preprint arXiv:2408.01800}.

\bibitem[{Zhang et~al.(2023)Zhang, Li, Zhang, Zhan, Wang, Zhou, and Qiu}]{zhang2023speechgpt}
Dong Zhang, Shimin Li, Xin Zhang, Jun Zhan, Pengyu Wang, Yaqian Zhou, and Xipeng Qiu. 2023.
\newblock Speechgpt: Empowering large language models with intrinsic cross-modal conversational abilities.
\newblock In \emph{The Conference on Empirical Methods in Natural Language Processing}, pages 15757--15773.

\bibitem[{Zhang et~al.(2024)Zhang, Chen, Hu, Han, Xu, Xu, Zhao, Sun, and Liu}]{zhang2024beyond}
Xinrong Zhang, Yingfa Chen, Shengding Hu, Xu~Han, Zihang Xu, Yuanwei Xu, Weilin Zhao, Maosong Sun, and Zhiyuan Liu. 2024.
\newblock Beyond the turn-based game: Enabling real-time conversations with duplex models.
\newblock In \emph{Proceedings of the Conference on Empirical Methods in Natural Language Processing}, pages 11543--11557.

\bibitem[{Zhou et~al.(2023)Zhou, Lu, Mishra, Brahma, Basu, Luan, Zhou, and Hou}]{zhou2023instruction}
Jeffrey Zhou, Tianjian Lu, Swaroop Mishra, Siddhartha Brahma, Sujoy Basu, Yi~Luan, Denny Zhou, and Le~Hou. 2023.
\newblock Instruction-following evaluation for large language models.
\newblock \emph{arXiv preprint arXiv:2311.07911}.

\bibitem[{Zhu et~al.(2024)Zhu, Liao, Zhang, Wang, Liu, and Wang}]{zhuvision}
Lianghui Zhu, Bencheng Liao, Qian Zhang, Xinlong Wang, Wenyu Liu, and Xinggang Wang. 2024.
\newblock Vision mamba: Efficient visual representation learning with bidirectional state space model.
\newblock In \emph{International Conference on Machine Learning}.

\end{thebibliography}

\appendix

\section{Prompt Template}
\label{sec:appendix}
We replace the "\{sentence\}" in Figure~\ref{fig.template} with predefined sentences as the prompt.
"\textbf{<speech>}" corresponds to the speech representation sequence.

\begin{figure*}[ht!]
\begin{center}
\begin{tcolorbox}[colback=gray!5!white,colframe=gray!75!black, width=0.68\textwidth]

<$|$user$|$> \\
\{sentence\} \\
<$|$beginofspeech$|$> \textbf{<speech>} <$|$endofspeech$|$> <$|$endofuser$|$> \\
<$|$assistant$|$> 

\end{tcolorbox}
\end{center}
\caption {\label{fig.template}Prompt template of DuplexMamba. \textbf{<speech>} corresponds to the speech representation sequence $\bm{S}$.}
\end{figure*}

\subsection{Sentences of ASR Prompt}
\label{sec:appendix_1}
\begin{enumerate}
  \item What does this audio say? Write it in lowercase without punctuation.
  \item Please convert this speech into text, all in lowercase and without punctuation.
  \item Generate the transcription for this audio without punctuation and keep it lowercase.
  \item Convert the spoken words into lowercase text without using any punctuation.
  \item Write the words you hear in this audio in lowercase, leaving out punctuation.
  \item Transcribe the speech in this audio to lowercase text with no punctuation.
\end{enumerate}

\subsection{Sentences of QA Prompt}
\label{sec:appendix_2}
\begin{enumerate}
  \item Please answer the questions in the user's input speech.
  \item Listen to this speech and provide an appropriate answer.
  \item Please respond to the questions asked in the audio.
  \item Based on this audio, provide a clear and concise answer.
  \item Respond to the query presented in this audio message.
  \item Please provide a response to the question in the speaker's voice.
  \item Respond to the audio's question with the appropriate answer.
\end{enumerate}

\section{Textual Labels}
\subsection{Incomplete Data}
\label{sec:appendix_3}
\begin{enumerate}
  \item It seems like your question got cut off. Could you please provide the full question so I can assist you better?
  \item It looks like your message got cut off. Could you please provide more details or restate your question? I'm here to help!
  \item It looks like your message was cut off. Could you please provide more details or complete your question? I'm here to help with whatever you need.
  \item It looks like your question isn't complete. Could you please provide a bit more detail or context so I can assist you better?
  \item It seems that your message got cut off. Could you please share your question again? I'm here to help!
  \item It looks like part of your message is missing. Could you kindly share the full question or clarify it further? I'd be happy to help!
\end{enumerate}

\subsection{Ignored Data}
\label{sec:appendix_4}
\begin{enumerate}
  \item I didn't get that, could you rephrase it?
  \item Sorry, could you explain that a bit more?
  \item Could you please elaborate on your question?
  \item Pardon me, but could you restate your question?
  \item I'm sorry, but I need a bit more context to understand.
  \item Apologies, I didn't quite catch what you meant.
  \item My apologies, I'm not sure I understood what you're asking.
  \item I'm sorry, could you clarify your question?
  \item I'm afraid I didn't fully understand your question.
  \item I'm not sure I follow—could you provide more details?
\end{enumerate}

\section{Configuration}
\label{sec.hyper}

\begin{table}[ht!]
\centering
\scalebox{0.78}{
\begin{tabular}{cccccc}
\toprule
\multirow{2}{*}{\textbf{Stage}} & \multirow{2}{*}{\textbf{GPUs}} & \multirow{2}{*}{\textbf{lr}} & \textbf{Warmup} & \multicolumn{1}{r}{\textbf{Max global}} & \multirow{2}{*}{\textbf{Epoch}} \\
&            &            & \textbf{steps}  & \textbf{batch size}                     &                         \\
\midrule
1                      & 6      & 2.5e-4              & 30000  & 3840                           & 7                       \\
2                      & 6      & 5e-5                & 20000  & 192                            & 2                       \\
3                      & 6      & 3e-5                & 15000  & 240                            & 3                       \\
4                      & 1      & 4e-5                & 30000  & 128                            & 2                       \\
\bottomrule
\end{tabular}}
\caption{Four-Stage training configuration.}
\label{tab:training_config}
\end{table}

The encoder of our trained ASR model consists of 12 layers of ConMamba blocks, while the decoder includes 6 layers of Mamba blocks. 
The model dimension is 512, and the Mamba state size is 16. 
Our Mamba-based language model, Mamba-2.8B, comprises 64 layers of Mamba blocks, with a model dimension of 2560 and a Mamba state size of 16.
The downsampling hyperparameter $k$ of the speech adapter is set to 5.

The four-stage training configuration of DuplexMamba is shown in Table~\ref{tab:training_config}. 
"GPUs" refers to the number of A100 GPUs used; "lr" and "Warmup steps" are hyperparameters required by the Noam Annealing scheduling strategy. 
We employ the dynamic batching method, and the "Max global batch size" corresponds to the theoretical maximum value for the batch size during training.

\begin{table}[ht!]
\centering
\scalebox{0.68}{
\begin{tabular}{lcc}
\toprule
\textbf{Model}    & \textbf{Speech Encoder} & \textbf{Language Model} \\
\midrule
DiVA              & Whisper-large-v3        & LLaMA-3-8B              \\
LLaMA-Omni        & Whisper-large-v3        & LLaMA-3.1-8B-Instruct   \\
Mini-Omni         & Whisper-small           & Qwen2-0.5B              \\
Mini-Omni2        & Whisper-small-v3        & Qwen2-0.5B              \\
Qwen2-Audio       & Whisper-large-v3        & Qwen-7B                 \\
VITA              & CNN+Transformer         & Mixtral-8x7B-v0.1       \\
Moshi             & Mimi                    & Helium-7B               \\
DuplexMamba(ours) & ConMamba-large          & Mamba-2.8B              \\
\bottomrule
\end{tabular}}
\caption{Model architecture of evaluated voice assistant models.}
\label{tab:model_architecture}
\end{table}

\end{document}